# Machine learning and emoji prediction: How much accuracy can MARBERT achieve?

Mohammed Q. Shormani, Ibb University, Yemen
shormani@ibbuniv.edu.ye/https://orcid.org/0000-0002-0138-4793
Ibrahim Abdulmalik Hassan, Ibb University, Yemen
ibrahim@ibbuniv.edu.ye
Muneef Y. Alshawsh, Ibb University, Yemen
alshawushmuneef@gmail.com

**Abstract**

This study investigates Machine Learning (ML) in the prediction of emojis in Arabic tweets employing the (state-of-the-art) MARBERT model. A corpus of 11379 CA tweets representing multiple Arabic colloquial dialects was collected from X.com via Python. A net dataset includes 8695 tweets, which were utilized for the analysis. These tweets were then classified into 14 categories, which were numerically encoded and used as labels. A preprocessing pipeline was designed as an interpretable baseline, allowing us to examine the relationship between lexical features and emoji categories. MARBERT was finetuned to predict emoji use from textual input. We evaluated the model performance in terms of precision, recall and F1-scores. Findings reveal that the model performed quite well with an overall accuracy 0.75. The study concludes that although the findings are promising, there is still a need for improving machine learning models including MARBERT, specifically for low-resource and multidialectal languages like Arabic.

## 1. Introduction

Machine learning (ML) is an artificial intelligence (AI) phenomenon in which machines/computers deeply learn to predict particular rules from given patterns. Differently put, ML is a branch of AI that enables computers to learn patterns from data and improve performance over time without explicit programming (see e.g. Sarker, 2021; Taye, 2023). It forms the backbone of Natural Language Processing (NLP), which focuses on enabling machines to understand, interpret, and generate human language (Goyal et al., 2018). By applying ML algorithms, NLP systems can perform tasks like sentiment analysis, language translation, and text summarization (Naithani et al., 2023). Early NLP relied on rule-based approaches, but ML shifted the field towards data-driven methods, allowing more accurate and scalable solutions. Deep learning, a subfield of ML, has further revolutionized NLP through models that capture complex language patterns and context. Today, ML-powered NLP is central to virtual assistants, chatbots, and AI-driven communication tools (see e.g. Goyal et al., 2018).

Additionally, the rapid growth of social media has introduced new forms of communication in which meaning is constructed through the integration of short textual messages and visual elements, most notably emojis (see e.g. Yus, 2025). With the increasing use of emojis across platforms such as X (Twitter), Facebook, WhatsApp, and Instagram, these pictorial symbols have become a de facto standard of online interaction, enabling users to express emotions, reactions, objects, and situational meanings in a concise and visually enriched manner. Consequently, emojis provide sentiment-rich data at a large scale, offering valuable opportunities for computational analysis and natural language processing research (Ahamad et al., 2025). Previous studies have investigated the meanings and functions of emojis primarily through textual contexts on social media platforms. However, despite their widespread use, relatively limited attention has been paid to emoji prediction, and even less, if any, attention has been given to emoji prediction in CA. this study, thus, attempts to bridge this gap.

As for studies that have addressed emoji prediction in social media, some studies employed BERT (e.g. Nusrat et al., 2011), some others MARBERT (e.g., Drous et al., 2025) and some others a combination (e.g., Abdul-Mageed & Elmadany, 2021). In this study, we employ MARBERT, which is a transformer-based language model, specifically designed for dialectical Arabic. It is built on the architecture of BERT and was trained on a very large corpus of Arabic text collected primarily from platforms like X.com (Abdul-Mageed & Elmadany, 2021). What makes MARBERT particularly powerful is its exposure to informal, noisy, and dialect-rich data, which allows it to handle CA much better than other models like BERT which are trained only on Standard Arabic (SA). It captures linguistic variation across dialects (e.g., Gulf, Egyptian, Levantine) and is widely used for tasks such as sentiment analysis, emoji prediction, dialect identification, and offensive language detection. In computational linguistics research, MARBERT represents a major advancement because it bridges the gap between formal Arabic NLP and real-world digital communication, making it highly suitable for studies like yours that focus on emojis and discourse in CA tweets see also (Abdul-Mageed & Elmadany, 2021).

Thus, the reminder of this article goes as follows. In section 1, we outline the theoretical foundations of the study, and review some related studies. In section 3, we present the study design, shedding light on data collection, tools, and methods. In section 4, we present the study results and discuss them. In section 5, we conclude the article.

## 2. Theoretical foundations

### 2.1. Machine learning: an overview

The very idea of machine learning emerged from the broader field of AI, which was formally established in the 1950s. One of the early pioneers of ML is Alan Turing whose seminal article (1950) *Can machine think?* has paved the way and opened windows to several developments in the field. Turing, in other words, laid the conceptual groundwork by asking whether machines could "think" and perform tasks that require human intelligence. In 1952, Arthur Samuel created one of the first self-learning programs, a checkers-playing program that improved its performance over time without explicit reprogramming (Samuel, 1953). This early work marked the beginning of machines being able to learn from data and experience, rather than relying solely on hard-coded instructions. The core conception of this has been the notion that if we understand how human brain processes, computes and interprets language, we will understand human intelligence, and if we put this "intelligence" in computers, the latter can then perform tasks similar to those performed by human (brains) (see e.g. Winograd, 1971; cf. also Chomsky, 1956).

During the 1960s and 1970s, machine learning research focused on symbolic approaches, attempting to encode knowledge and reasoning rules explicitly, using methods such as decision trees and early neural networks. The 1980s and 1990s saw a resurgence in machine learning driven by improvements in algorithms and computational power. The backpropagation algorithm revitalized interest in artificial neural networks, enabling multi-layered networks to learn complex patterns. At the same time, statistical approaches such as Bayesian methods, support vector machines, and clustering algorithms gained popularity (see e.g. Vaswani et al., 2017). Researchers began to shift from symbolic AI to data-driven approaches, recognizing that machines could discover patterns from large datasets without requiring exhaustive human programming. This period laid the foundation for modern supervised and unsupervised learning methods. In the 2000s and 2010s, the field experienced exponential growth due to the rise of "big data," faster processors, and advances in deep learning. Neural networks such as Convolutional Neural Networks (CNNs) and Recurrent Neural Networks (RNNs) allowed breakthroughs in image recognition, NLP, and speech understanding (Vaswani et al., 2017; Sohail et al., 2023; Shormani, 2024). These CNNs and RNNs enable models such as MARBERT to learn linguistic patterns the same way a child acquires/learns linguistic patterns via complex mental computational systems (see e.g. Chomsky, 1957, 2013; Shormani, 2014, 2025a).

Big companies like Google, Microsoft, and Facebook began investing heavily in ML research, and open-source frameworks like TensorFlow and PyTorch made these technologies widely accessible (see e.g. McShane & Nirenburg, 2021). Today, ML continues to evolve rapidly, influencing fields ranging from healthcare and finance to autonomous systems and AI-driven creative tools (Shormani, 2025b).

### 2.1.1. MARBERT

Recent advances in NLP have led to a huge change with the implementation of neural networking algorithms (NNAs), which result in several sophisticated models, largely driven by transformer-based architectures such as BERT. These models rely on self-attention mechanisms that allow them to capture contextual relationships between words in a sentence, enabling deeper semantic understanding compared to traditional models. Vaswani et al.'s (2017) work has significantly affected NLP field, specifically introducing Transformer architecture and BERT. MARBERT was built based on BERT architecture, having 12-layered transformers, and 163 million trainable parameters (Abdul-Mageed et al., 2022). MARBERT has been used by several studies and scholars, having outperformed several models including XLM-R Large (Conneau et al., 2020), AraGPT2 (Antoun et al., 2020), and CAMeLBERT (Abdul-Mageed et al., 2022). MARBERT is designed for CA dialects, unlike BERT which can work for SA, the former having been trained on large-scale social media outlets' data (see also Drous et al., 2025). It is specifically effective for modeling dialectal and informal Arabic, making it well-suited for analyzing CA tweets in our study.

Employing MARBERT, this study draws on principles of statistical learning, particularly supervised classification methods used in NLP. Logistic Regression was also utilized, which operates on feature representations like TF-IDF, encoding the importance of words in a corpus. These models assume that linguistic patterns can be captured through statistical associations between features (words) and labels (emoji classes). While less powerful than deep learning models, Logistic Regression provides interpretability by identifying which lexical features are most predictive of specific outcomes (see e.g. Levy & O'Malley, 2020). This aligns with the view that language exhibits probabilistic regularities that can be learned from data and used for prediction tasks (see e.g. Murphy, 2012).

## 2.2. Emojis

Emojis can be understood as paralinguistic features that complement textual communication by encoding emotional, interpersonal, and contextual information (Yus, 2025). Unlike purely textual elements, emojis function as visual cues that enhance or modify meaning, often signaling affective states such as happiness, sadness, or sarcasm (Ullah et al., 2025). Prior research (see e.g. Yus, 2014) demonstrates that emojis are not randomly used but are systematically associated with linguistic patterns, making them predictable from textual input. This perspective aligns with broader theories of multimodal communication, where meaning is distributed across different semiotic resources. In computational spheres, emojis can therefore be modeled as labels that reflect underlying semantic and emotional structures in language (Yus, 2025). From an NLP perspective, investigating emojis is still relatively unexplored with some studies focusing on sentiment analysis (see e.g. Drous1et al., 2025). There are relatively few studies that have addressed their prediction in socal media outlets (see e.g. Zhao et al., 2018; Barbieri et al., 2018; Nustrat et al., 2023).

As for our study, the question is: Can emojis be predicted, and specifically in CA? And if so, to what extent? Emoji prediction is theoretically framed as a multi-class classification problem, where the objective is to map textual input to one of several predefined emoji categories. This task has been explored in prior work, including studies using deep learning models such as BERT (e.g., Nusrat et al., 2023) and feature-based approaches incorporating contextual metadata (Venkit et al., 2021). These studies demonstrate that both linguistic content and contextual signals contribute to emoji usage. Within this framework, the present study extends previous work by focusing on CA and employing a fine-grained categorization of emojis, thereby contributing to a more detailed understanding of emoji prediction in low-resource like CA and dialectally rich contexts such as X.com.

### 2.3. Colloquial Arabic

CA is often referred to as *ʕammiyyah* (عامية), the everyday spoken dialects used by native speakers across the Arab world (see e.g. Bassiouney, 2009). Unlike SA, which is primarily used in formal writing, news broadcasts, and religious contexts, CA is the natural, fluid language of daily life including conversations at home, in cafes, on the street, and in popular media like TV series and songs (see e.g. Holes, 2004). However, a key feature is its vast regional diversity. For instance, a speaker from Cairo might say إزيك *ʔizzayyak* 'How are you?' while someone in Levantine says كيفك *kifak*. These dialects differ significantly in pronunciation, vocabulary, and even grammar, sometimes to the point of mutual unintelligibility across distant regions, such as between a Moroccan and an Iraqi dialect.

Thus, CA refers to a spectrum of living dialects that constantly evolve and borrow from each other, specifically through media (and migration). Importantly, while SA remains the formal standard, using CA in everyday settings signals cultural integration and builds genuine human connection, as it is the mother tongue for over 400 million people. Ultimately, mastering a local dialect opens the door to the humor, poetry, and warmth of everyday Arab life that formal Arabic alone cannot fully capture.

### 2.4. Web scraping

Social media outlets such as Facebook, X.com, Telegram are now considered substantial tools, representing discursive views. Their data have been a major source for scholars, researchers and linguists, computer scientists, and sentiment analysts for the valuable data they provide. One of the widely used methods for acquiring data is web scraping or crawling (Khder, 2021). Web scraping is an effective approach for automatically generating online content (see e.g., Hajba, 2018). Many researchers use web scraping to create content or gather feedback, enhancing the accuracy of marketing strategies, supporting resource development, and driving business growth (Hajba, 2018; Thomas et al., 2019). Web scraping, known also as web data extraction, is designed to collect significant data from various online sources and compile them for new applications. A web scraper retrieves information from websites and is commonly used for purposes such as order processing, data mining, price monitoring and comparison, competitor review tracking, sentiment analysis, real estate listing aggregation, weather data monitoring, site change detection, online reputation tracking, and web mashups (Hajba, 2018; Hernandez-Suarez et al., 2018).

*Tweepy* is a Python library which is made use of to extract data from X.com. One of *Tweepy* strengths lies in its support for both restful and streaming APIs (Dongo et al., 2021). The restful API allows users to request specific data such as tweets from a certain user, follower lists, or tweet details. The streaming API, on the other hand, provide real-time access to tweets that match certain filters or track certain keywords (see e.g. Hajba, 2018). This real-time capability is particularly useful for projects involving sentiment analysis, event tracking, or live data monitoring. *Tweepy* is widely used in research, data science, and machine learning projects, specifically in fields like social media analytics, sentiment analysis, and digital marketing (Shormani & Alenezi, 2026). By combining *Tweepy* with Python data analysis and visualization libraries, such as Pandas and Matplotlib, developers can collect, clean, and analyze large volumes of Twitter data efficiently (Dongo et al., 2021). Although recent changes to the Twitter API and access policies require more stringent authentication and sometimes paid access, *Tweepy* remains a powerful tool for connecting Python applications to Twitter ecosystems, which we utilized in our study.

### 2.5. Previous studies

To the best of our knowledge, no previous study has tackled this topic. Perhaps this is also true concerning using MARBERT in this juncture. However, several studies have addressed emojis prediction in social media outlets. For example, Venkit et al. (2021) investigate emoji prediction by incorporating both textual and non-textual features of tweets. The study extends beyond traditional text-based approaches by integrating additional contextual signals, including sentiment polarity, hashtags, and the application source used to post tweets. Treating emoji prediction as a multi-class classification task, the authors demonstrate that combining these features leads to improved performance compared to models relying solely on textual input. Their findings highlight that hashtags and sentiment provide strong cues about the intended meaning of a tweet, while the application source reflects user behavior patterns that can influence emoji usage. The study concludes that leveraging multiple sources of information offers a more comprehensive understanding of emoji prediction and enhances model effectiveness in social media analysis.

Abdul-Mageed and Elmadany (2021) introduces ARBERT and MARBERT, two deep bidirectional transformer models, specifically designed for Arabic NLP. Both models are based on the architecture of BERT, but differ in their training data and intended use. ARBERT is trained primarily on SA texts, making it suitable for SA tasks, while MARBERT is trained on large-scale Arabic social media data, enabling it to effectively handle dialectal and informal varieties of Arabic. The authors evaluate both models across a range of NLP tasks, including sentiment analysis, dialect identification, and offensive language detection, demonstrating that they significantly outperform previous models.

Aboutaib et al. (2025) present a comparative analysis of transformer-based models for the prediction of punctuation in Arabic texts. The authors evaluate several pretrained transformer architectures, including models based on BERT and its Arabic adaptations, to determine their effectiveness in restoring punctuation in unpunctuated Arabic data. Treating punctuation prediction as a sequence labeling task, the study demonstrates that transformer models significantly outperform traditional approaches due to their ability to capture contextual dependencies within text. The results highlight the importance of contextual embeddings and show that model performance varies depending on the type of Arabic data (e.g. standard vs. dialectical). Overall, the study confirms the effectiveness of transformer-based approaches for Arabic language processing tasks and underscores their potential for improving downstream applications such as readability, speech processing, and text understanding.

Another study was conducted by Barbieri et al. (2017) examining whether emojis can be predicted from textual content in social media, specifically Twitter. They formulate emoji prediction as a multi-class classification task and employ neural network models, including Bidirectional Long Short-Term Memory (BiLSTM) architectures, to capture contextual and semantic features of tweets. Their results show that emojis are highly predictable from text, indicating a strong correlation between linguistic patterns and emoji usage. The study also demonstrates that neural models outperform traditional approaches, highlighting the importance of contextual representation in understanding informal digital communication. findings suggest that emojis function as meaningful extensions of text rather than random additions, contributing valuable information for natural language processing tasks such as sentiment analysis.

Perhaps the most related study to ours is conducted by Nusrat et al. (2023) in which they investigate emoji prediction in Twitter data using BERT. Treating the task as a multi-class classification problem, they finetune a pretrained BERT model to capture contextual and semantic features of tweets. Their results show that BERT-based models outperform traditional machine learning approaches, demonstrating the effectiveness of contextual embeddings for handling short and informal text. The study concludes that transformer-based models significantly enhance emoji prediction performance and can support related tasks such as sentiment analysis and social media understanding. This study differs from ours in many and several aspects including i) they used BERT while ours employs MARBERT, and ii) they tackled English tweets, but our study involves CA tweets.

Thus, building on this line of research, the present study investigates emoji prediction in CA tweets using an corpus-based approach, and utilizing MARBERT. By focusing on a multidialectal Arabic dataset, this study addresses a significant gap in the literature, as most previous work has concentrated on English or other high-resource languages such as SA. In addition, this study expands the scope of emoji analysis by adopting a fine-grained categorization of emojis into multiple functional classes and by employing advanced machine learning techniques to model their usage. Through this approach, the study aims to contribute to a deeper understanding of how emojis function in Arabic digital communication and how they can be effectively predicted using MARBERT, endeavoring to answer the following questions:

1. Can emojis be predicted in X.com tweets?
2. To what extent can emojis in CA tweets be accurately predicted using MARBERT?
3. What are the most problematic challenges of this prediction?

## 3. Study design

### 3.1. Datasets

We collected 11379 tweets utilizing Python 9.3.11 via API2. These tweets were saved in a CSV file. We found 2684 tweets without emojis, which were excluded. We divided the remaining tweets, i.e. 8695 with emojis, into 2 datasets: dataset 1 contains 4000, and dataset 2 includes 4695 tweets.

We used dataset 1 as a training dataset to finetune emojis use and occurrences in the corpus, and dataset 2 for testing the model. We identified 14 most recurrent types of emojis commonly used in Arabic social media. These are summarized in Table 1.

**Table 1: Emoji category and examples**

| Category | Emoji |
|---|---|
| Happiness | 😂, 😄, 😁 |
| Love | 😍, ❤️, 💕 |
| Sadness | 😢, 😭, 😥 |
| Anger | 😡, 😠. 🤬 |
| Fear | 😨, 😱, 🥶 |
| Surprise | 😲, 🤯, 😳 |
| Prayer | 🙏, 🤲, 📿 |
| Calmness | 😌, 😊, 😇 |
| Confusion | 🤔, 🧐, 🤨 |
| Confidence | 😎, 💪, 😏 |
| Sarcasm | 🤪, 😜, 🤣 |
| Embarrassment | 😳, 😅, 🤦‍♀️ |
| Weariness | 😫, 😩, 😰 |
| Stress | 🤯, 🥴, 😴 |

We then import MARBERT on Colab, and developed the Python script code as displayed in Fig 1, through which we performed the training process.

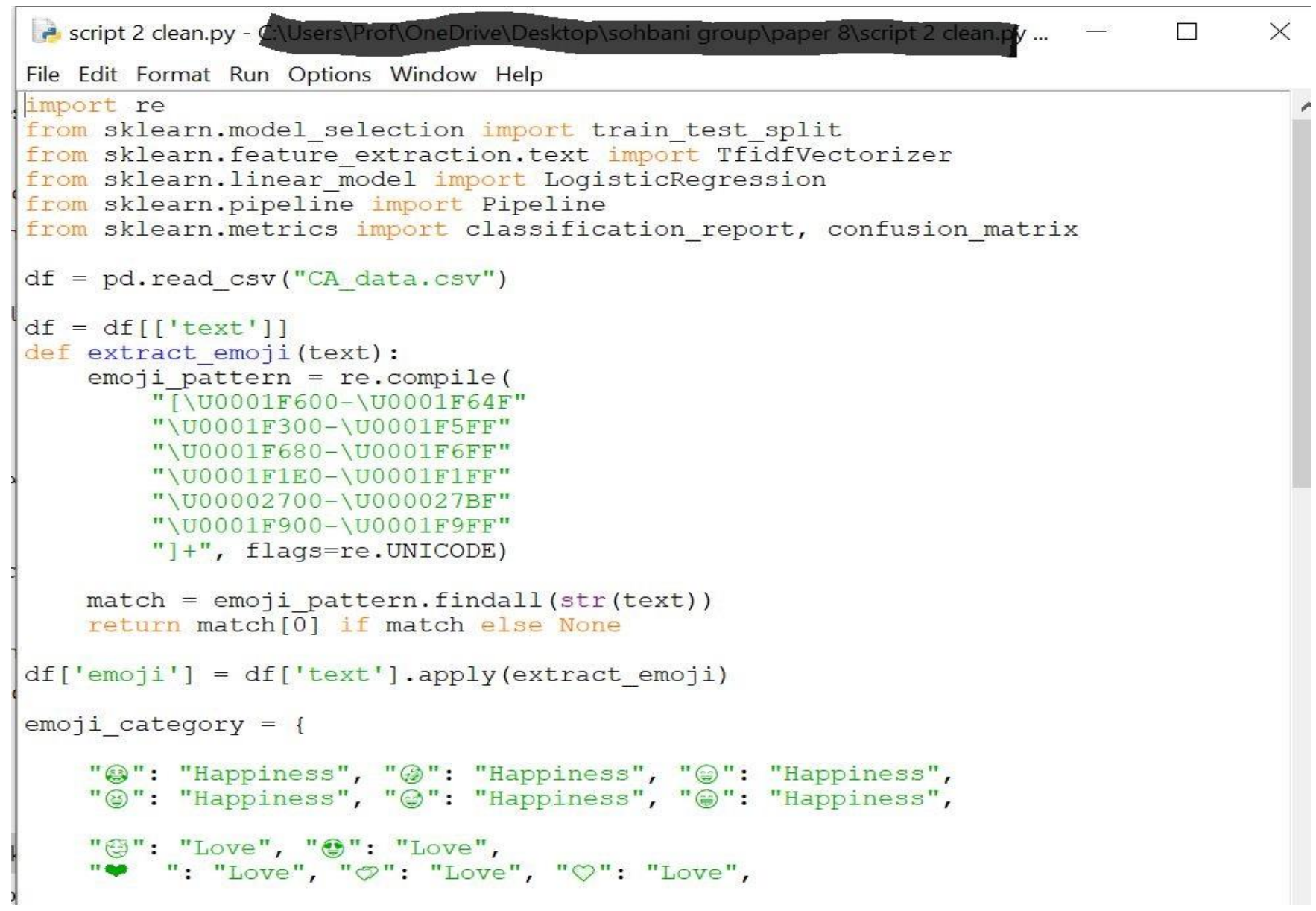

```
script 2 clean.py - C:\Users\Prof\OneDrive\Desktop\sohbani group\paper 8\script 2 clean.py ...
File  Edit  Format  Run  Options  Window  Help
import re
from sklearn.model_selection import train_test_split
from sklearn.feature_extraction.text import TfidfVectorizer
from sklearn.linear_model import LogisticRegression
from sklearn.pipeline import Pipeline
from sklearn.metrics import classification_report, confusion_matrix

df = pd.read_csv("CA_data.csv")

df = df[['text']]
def extract_emoji(text):
    emoji_pattern = re.compile(
        "[\U0001F600-\U0001F64F"
        "\U0001F300-\U0001F5FF"
        "\U0001F680-\U0001F6FF"
        "\U0001F1E0-\U0001F1FF"
        "\U00002700-\U000027BF"
        "\U0001F900-\U0001F9FF"
        "]+", flags=re.UNICODE)

    match = emoji_pattern.findall(str(text))
    return match[0] if match else None

df['emoji'] = df['text'].apply(extract_emoji)

emoji_category = {

    "😂": "Happiness", "🤣": "Happiness", "😄": "Happiness",
    "😆": "Happiness", "😅": "Happiness", "😁": "Happiness",

    "🥰": "Love", "😍": "Love",
    "❤  ": "Love", "💞": "Love", "♡": "Love",
```

*Fig 1: Part of Python training script*

We ran the Python code on (Google) Colab as it provides a cloud-based platform that supports Python execution without requiring local installation or configuration, which facilitates accessibility and reproducibility. It also offers built-in integration with essential libraries such as *pandas* and *scikit-learn*, as well as support for GPU acceleration when needed. Additionally, Colab enables easy data handling through integration with cloud storage and simplifies experimentation by allowing code execution in modular cells with several useful parameters. These features make it particularly suitable for processing large datasets and conducting iterative machine learning experiments efficiently. Colab also integrates seamlessly with GoogleDrive, enabling effective storage, organization, and retrieval of notebooks and related project files (cf. also Kanani & Padole, 2019).

### 3.2. Data screening

After collected the data, saving them as CSV files, there were many noise elements that have been screened and refined before training. These things include inconsistent, noise and irrelevant characters such as 3 dots '…', symbols as '&', '*', '%', hashtags, repeated letters like ي as in كتييير , و as in اوووووي, and ا as in عاااااايز. Other irrelevant elements include symbols like '@' with account names (e.g. @XXX, XXX stands for personal names, we avoided spelling them out for user privacy). There were also unrecognized emojis like ✳, 🔕, 🏚. All these noise elements were removed utilizing Python, except repeated or longed letters which were removed manually. These inconsistent and irrelevant things were removed from both datasets before training. Our aim was to make the datasets clear from these noise elements that may affect training and testing processes.

### 3.2. Pipeline

A supervised machine learning Pipeline was implemented to predict emoji use in CA tweets. Emojis were first categories and mapped into 14 categories (cf. Table 1). These categories were numerically encoded and used as labels. A TF-IDF vectorizer combined with logistic regression was employed for classification. Model performance was evaluated using precision, recall, F1-score, and confusion matrices.

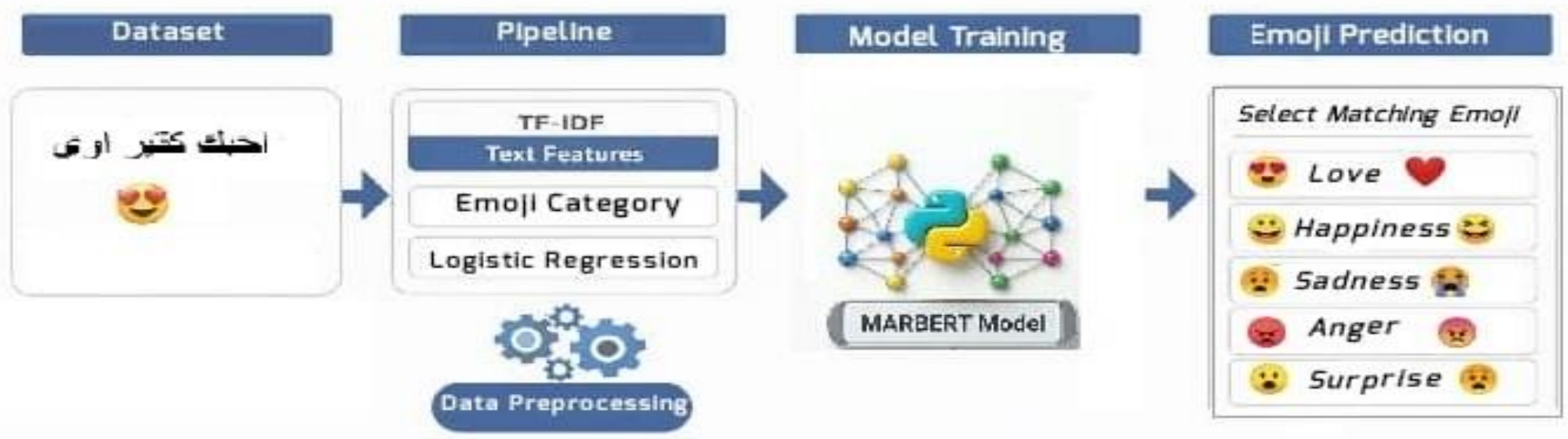

*Fig. 2: MARBERT workflow*

Fig 2 presents the supervised machine learning Pipeline for emoji-based text classification from a dataset. It shows how the whole workflow takes place. First the dataset is uploaded into the model, which consists of textual tweets and emojis. Second, the Pipeline comes to play, where TF-IDF extracts text features and mapped them to emoji type(s) (in the form of category see Table 1, where one or more emojis may be predicted). Put differently, tweets are processed to map emojis using regular expressions, which are then mapped into predefined classes and encoded numerically. The dataset is split into training and testing sets, and text features are extracted using TF-IDF. Third, a Logistic Regression classifier is trained to classify emoji categories, given the emoji classes predefined in the Python script, and model performance is evaluated using standard metrics.

Additionally, tokenization of tweets and emojis takes the form of mapping and representing of these elements as numerical vectors in MARBERT architecture. The vectors help the model process and compute matching tweets to emojis in the form of semantic representation, which in turn helps the model deeply learn the relationship between texts and emojis. Given that MARBERT has been pretrained on large CA datasets, if we train it on our data, its neural mechanism allows it to capture text and map it to emojis. After training process complete, the model becomes ready to be tested on the testing data, in which it predicts which emoji occurs with which text. After training MARBERT, we evaluated its performance employing evaluating metrics. This is tackled in the following section.

### 3.3. Evaluating metrics

For evaluating the model performance, we used the following metric formula (cf. e.g. Nusrat et al., 2023).

$$F1 - score \ = 2\ x\ \frac{(precision\ x\ recall)}{(precision + recall}$$

## 4. Results and discussion

In this section, we tabulate the results and analyze them. Table 2 summarizes the results in terms of class, precision, recall, F1-score and support. The latter points to the instances of the class of emojis in the testing data.

**Table 2: MARBERT performance**

| Class | Precision | Recall | F1-score | Support |
|---|---|---|---|---|
| 0 | 0.66 | 0.77 | 0.67 | 180 |
| 1 | 0.84 | 0.87 | 0.86 | 1393 |
| 2 | 0.73 | 0.76 | 0.74 | 278 |
| 3 | 0.51 | 0.68 | 0.58 | 144 |
| 4 | 0.81 | 0.83 | 0.85 | 272 |
| 5 | 0.41 | 0.65 | 0.50 | 217 |
| 6 | 0.42 | 0.74 | 0.54 | 123 |
| 7 | 0.70 | 0.78 | 0.74 | 109 |
| 8 | 0.39 | 0.62 | 0.48 | 60 |
| 9 | 0.48 | 0.71 | 0.57 | 35 |
| 10 | 0.47 | 0.84 | 0.60 | 119 |
| 11 | 0.42 | 0.73 | 0.54 | 115 |
| 12 | 0.59 | 0.84 | 0.69 | 111 |
| 13 | 0.71 | 0.58 | 0.63 | 47 |

Considering Table 2, it seems that the model performs relatively well. The results of the finetuned MARBERT model for emoji prediction in CA tweets show a generally strong performance across classes, reflecting the strengths of transformer-based contextual modeling between texts and emojis. The best-performing categories are those with either relatively large training support or highly salient lexical-emotional cues. In particular, Class 1 demonstrates the highest and most stable performance, with precision 0.84, recall 0.87, and F1-score 0.86. This strong performance is expected given its large support 1393 instances, which allows the model to learn robust contextual patterns associated with affectionate and positive expressions. Similarly, classes 4 and 7 also show strong or near-strong performance, with F1-scores of 0.85 and 0.74, respectively. However, mid-range classes include 2, 10, and 11 showing moderate performance, with F1-scores ranging between approximately 0.54 and 0.74. These classes are typically broader in meaning and more context-dependent, which increases the likelihood of overlap with other emotional classes. The lowest-performing categories include classes 5, 6, 8, and 9 with F1-scores ranging from approximately 0.48 to 0.57.

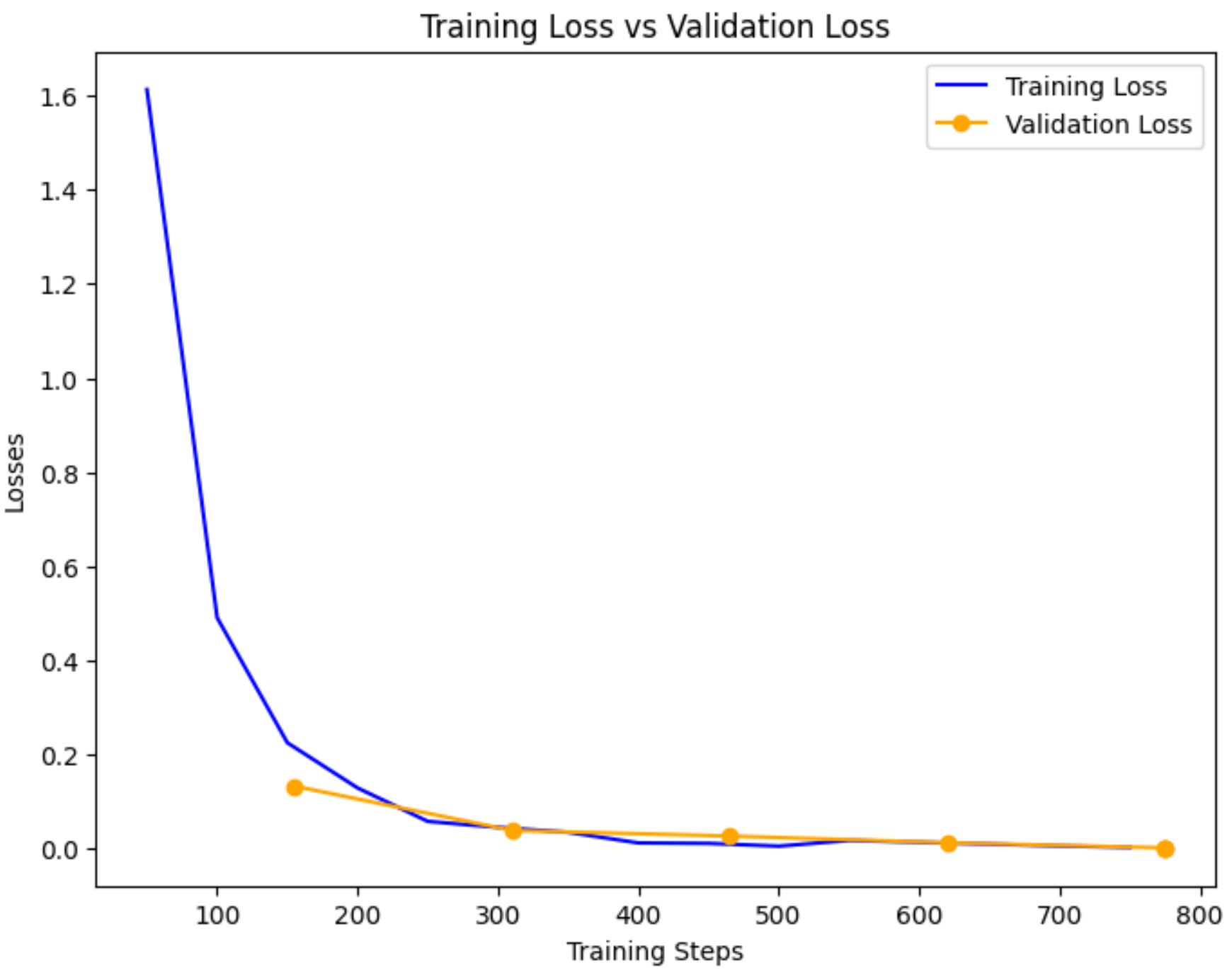

***Fig 3: MAREBERT Training Loss vs Validation Loss***

Fig 3 shows the gradual evolution of training and validation losses across training steps, illustrating a clear and stable optimization process. The training loss decreases sharply at the early stages of training, dropping from a relatively high initial value, viz. ≈1.6 to below 0.5 within the first ~100 steps. This rapid decline indicates that the model quickly captures the most salient patterns in the training data, reflecting efficient initial learning and effective gradient-based optimization. After the initial phase, the training loss continues to decrease more gradually, approaching near-zero values by approximately 400–800 steps. This slower rate of improvement suggests that the model transitions from learning coarse-grained patterns to finetuning more subtle aspects of the data representation. The smooth and monotonic decline, with no visible oscillations, further indicates stable convergence without signs of optimization instability. The validation loss follows a similar decreasing trend, albeit with fewer evaluation points. It starts at a relatively low value compared to the initial training loss and continues to decrease steadily as training progresses. Importantly, the validation curve closely tracks the training curve throughout most of the training process. By the later stages, both losses converge to very small values, indicating strong agreement between training performance and generalization performance.

Note that there is no clear evidence of overfitting in the presented curves. The absence of divergence between training and validation losses, together with their simultaneous decline towards near-zero values, suggests that the model maintains good generalization capability. The small gap observed in earlier steps quickly diminishes, reinforcing the conclusion that the model does not memorize the training data at the expense of validation performance. This is also supported by Training and Validation Dynamics MARBERT undergoes in this study as depicted in Table 3.

**Table 3: Training and validation dynamics of MARBERT**

| Step | Epoch | Training loss | Validation loss | Learning rate |
|---|---|---|---|---|
| 50 | 0.32 | 1.613 | — | 4.68e-05 |
| 100 | 0.65 | 0.491 | — | 4.36e-05 |
| 150 | 0.97 | 0.225 | — | 4.04e-05 |
| 155 | 1.00 | — | 0.133 | — |
| 200 | 1.29 | 0.129 | — | 3.72e-05 |
| 250 | 1.61 | 0.058 | — | 3.39e-05 |
| 300 | 1.94 | 0.046 | — | 3.07e-05 |
| 310 | 2.00 | — | 0.039 | — |
| 350 | 2.26 | 0.035 | — | 2.75e-05 |
| 400 | 2.58 | 0.013 | — | 2.43e-05 |

Table 3 shows that the training process was monitored through step-wise logging of loss, gradient norm, and learning rate. The training loss decreased significantly from 1.61 at step 50 to 0.01 at step 400, indicating rapid and effective learning. Similarly, validation loss dropped from 0.133 at the end of the first epoch to 0.039 at the second epoch, demonstrating strong generalization performance. The gradual decay in learning rate and reduction in gradient norms further confirm stable convergence of the model.

To recapitulate, the learning curves indicate a well-behaved training process characterized by fast initial convergence, smooth optimization dynamics, and strong generalization. The close alignment between training and validation losses throughout training suggests that the model has successfully learned representations that transfer effectively to unseen data.

### 4.1. Emoji prediction in CA via MAREBRT

We turn now to exemplify our model prediction. Fig 4 presents a sample of prediction made by the model, both correct and incorrect predictions.

Text: احبك كتير اوي
Prediction: 😍
Actual: 😍

Text: ذابحني ذبح
Prediction: 😩
Actual: 😩

Text: مزاج اليوم مابي أتحرك من مكاني ، الله يخلصنا بس
Prediction: 😴
Actual: 😴

Text: انتو كلمه السر فى المباراه دى ان شاء الله
Prediction: 💪
Actual: 💪

Text: Text: ماتلاحظين انك دايم تغردين بالموضوع الغير مناسب بالوقت الغلط
Prediction: 😂
Actual: 🤣

Text: يويلي نموت فيك
Prediction: 😂
Actual: ❤️

Text: مبروك عقبالي
Prediction: 😔
Actual: 😍

***Fig 4: Model prediction vs. actual use***

Fig 4 shows some examples of MARBERT prediction and actual use of emojis in our corpus. The first-four examples depict where the model performs very accurately. Expressions like " احبك كتير اوي" *ʔaḥibbak kitir ʔawi* 'I love you so much, "ذابحني ذبح" *dābiḥnī dahbḥ:* 'killing me indeed', and "مزاج اليوم مابي أتحرك..." *mizāj al-yawm mā abī ataḥarrak*... 'today I'm not in the mood to move…' are all correctly mapped to their corresponding emojis 😍, 😩, 😴. These cases share a common feature. Similarly, the sentence "انتو كلمه السر..." *ʔintu kalimat al-sirr* ...'you are the key to winning…' is correctly classified, which provides strong support for the finetuning of the model. However, the last three examples highlight where the model loses accuracy. For example, in "ماتلاحظين انك دايم تغردين..." *mā tulāḥiẓīn innik dāyim teɣarridīn*... 'don't you notice that you are always talking ….', the model predicts 😂 instead of 🤣, which, while close, still reflects a failure to capture nuanced pragmatic force, such as intensity or tone. More critically, in "يويلي نموت فيك" *yā waylī namūt fīk* 'Oh my Goodness, I adore you', the model outputs 😂 whereas the actual emoji is ❤️, indicating a misprediction of exaggerated humor rather than love. The most striking mismatch appears in "مبروك عقبالي 😔" *mabrouk ʕuqbāli* 'Heartfelt congratulations, hope I am next', where the model predicts 😔 but the actual emoji here is 😍.

These incorrect predictions can be ascribed to confusion, which in turn can be attributed to factors including the fact that some emojis have similar semantic and pragmatic denotations, i.e. some emojis may belong to several categories such as weariness and stress, for example. This makes MARBERT confuse some emojis for different classes. Although a confusion matrix was examined during evaluation, the results indicate that most misclassifications occur between semantically related emoji categories, highlighting the inherent ambiguity of emotional expression in social media text.

## 5. Conclusions and implications

To conclude, this study tackles a very important topic investigating ML in predicting emojis in CA tweets employing MARBERT model. 4695 tweets were used for testing the model after training it on 4000 tweets with emojis. Our evaluation of the model results in relatively well precisions, recalls and F1-scores, the highest of which are 0.84, 0.87, and 0.86, respectively, and an overall accuracy 75%. Thus, several conclusions could be drawn from our study: i) the results of precision, recall and F1-score and the overall all accuracy are promising, ii) this implies that emojis are predictable, which is in line with studies such as (Barbieri et al., 2017; Zhao et al., 2018; Nusrat et al., 2023), iii) MARBERT testing process results in 3203 tweets out of total 4695, i.e. about 68% of the total number of testing tweets. This is also another encouraging finding of this study, and; however, iv) that the overall accuracy is 75% suggests that MARBERT can be developed further for better results. This is also to ascertain that there is still a dire need for improving machine learning models, specifically for low-resource and multidialectal languages like Arabic, v) in our study, the MARBERT model seems to outperform BERT (Nusrat et al., 2023), XLM-R Large (Conneau et al., 2020), AraGPT2 (Antoun et al., 2020), and CAMeLBERT (Abdul-Mageed et al., 2022), and vi) we used emoji classes (see Table 2), and not individual emojis.

Additionally, MARBERT achieved an overall accuracy of 75%, with macro-averaged precision, recall, and F1-score of 0.58, 0.74, and 0.64, respective. The higher recall indicates that the model effectively identifies emoji classes, while the relatively lower precision suggests some confusion among semantically similar classes. This suggest that, though the accuracy is relatively strong, there is a need for improving machine learning models, specifically for low-resource and multidialectal languages like Arabic. These findings support the complementary role of deep contextual models and suggest that future improvements may benefit from data augmentation for low-resource categories and finer-grained annotation schemes that reduce semantic overlap between emotionally adjacent classes (cf. e.g. Mahardhika et al., 2025).

Despite the promising results obtained, this study has several limitations: i) the classification process was conducted by the model, which may not be strong enough. A future study could tackle data classified by human, ii) the size of the datasets for both training and testing the model. 8695 may not be enough for robust training as well as testing. A further study could be conducted on large datasets, say 50000-100000 tweets, and iii) involving one platform may not be enough to identify the diversity of emoji use. A broader study could involve more than one platform like X.com, Facebook, Instigram.